\documentclass[runningheads]{llncs}
\usepackage[T1]{fontenc}
\usepackage{graphicx}
\usepackage{amsmath}
\usepackage{booktabs}
\usepackage{longtable}
\usepackage{multirow}
\usepackage{enumitem}
\usepackage{array}
\usepackage{amssymb}
\usepackage{amstext}
\usepackage{multicol}
\usepackage{pslatex}
\usepackage{pifont}
\usepackage{tikz}
\usetikzlibrary{shapes}
\usepackage{pifont}
\usepackage{bbding}
\begin{document}
\title{TadML: A fast temporal action detection with Mechanics-MLP}
%
%
\author{Bowen Deng\inst{1,2} \and
Shuangliang Zhao\inst{2}\and
Dongchang Liu\inst{1}}
\authorrunning{Bowen Deng et al.}
%
\institute{Institute of Automation, Chinese Academy of Sciences, Beijing, 100190, China 
\email{dongchang.liu@ia.ac.cn}\\
\and
Guangxi University, Nanning, 69121, China\\
\email{szhao@gxu.edu.cn}}
\maketitle              
\begin{abstract}
Temporal Action Detection (TAD) involves identifying action categories and their respective start and end frames in lengthy untrimmed videos, with current models utilizing both RGB and optical flow streams that require manual intervention, add computational complexity, and consume time. Moreover, two-stage approaches prioritizing proposal generation in the ini-tial stage result in a substantial reduction in inference speed. To address this, we propose a single-stage anchor-free method that solely utilizes the RGB stream and incorporates a novel Newtonian Mechanics-MLP architec-ture. Our model achieves comparable accuracy to existing state-of-the-art models but with significantly faster inference speeds, clocking in at an av-erage of 4.44 videos per second on THUMOS14. Our approach showcases the potential of MLP in downstream tasks like TAD. The source code is available at https://github.com/BonedDeng/TadML.

\keywords{Temporal action dcetection  \and MLP-like \and RGB and optical flow \and Real time \and Anchor free.}
\end{abstract}
\section{Introduction}
As videos become ubiquitous in the wake of advances in mobile communication and the internet, video understanding has become increasingly important in both academia and industry. In particular, temporal action detection, detecting categories, start and end timestamps of human actions in untrimmed footage, has diverse applications in areas such as human-computer interaction, video surveillance, and intelligent security\cite{ref1}. In the past, numerous TAD frameworks employed complex pipelines. Some earlier methods even utilized manually crafted features, including color and texture features of each frame, for video action classification and detection. Currently, research on temporal action detection (TAD) has shifted towards utilizing deep models that combine raw RGB streams and optical flow, emerging as the mainstream and potential approach. RGB frames contain vital shape and spatial information of videos, which are necessary for Temporal Action Detection (TAD). Most TAD research also use optical flow, a two-dimensional velocity field that captures action information and three-dimensional scene structure of the observed object. However, fusing RGB and optical flow data requires time-consuming conversion, computations, and resources.
\begin{figure}
\centering
\includegraphics[width=0.45\textwidth]{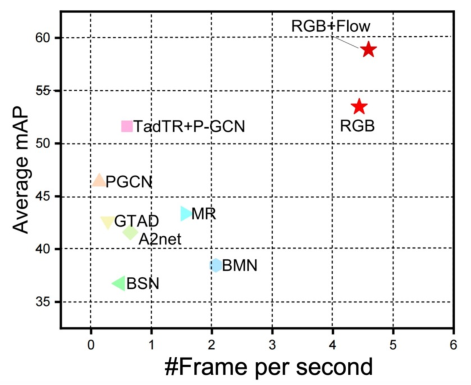}
\caption{ Comparing to the performance (average chart) and speed of the latest time action detection model on THUMOS 14. Our method shows advanced performance and very fast speed when using RGB stream.} \label{fig1}
\end{figure}
TAD has two objectives: predicting action categories based on available information and their corresponding timestamps in the video. While TAD shares similarities with object detection, it is focused on detecting actions in the time domain, unlike object detection that identifies positions in the spatial domain, leading TAD methods to draw inspiration from object detection research. TAD models can be categorized into one-stage and two-stage models based on their network structure. Two-stage frameworks generate proposals with high recall, which are then classified to predict corresponding labels, but their inference speed is slower and incurs higher computing costs than one-stage frameworks. One-stage frameworks simultaneously generate the start and end frames of each action and their corresponding labels in a single step, making them more efficient for real-time applications. 

Based on the anchor structure, previous research can be categorized into three groups: (a) action-guided methods, such as BSN\cite{ref2}, (b) anchor-based methods, such as BMN\cite{ref3}, and (c) anchor-free methods, such as AFSD\cite{ref4}. Methods that employ anchors not only exhibit high time complexities, namely $(T^2)$ and $(c*t)$, but also require numerous hyperparameters to be fine-tuned, including the scale and quantity of anchors, as well as the computational cost of IOU thresholds\cite{ref5}. Drawing inspiration from anchor-free models in target detection research, anchor-free methods have emerged as the mainstream approach and demonstrated significant potential in TAD.
\begin{figure}
\centering
\includegraphics[width=0.4\textwidth]{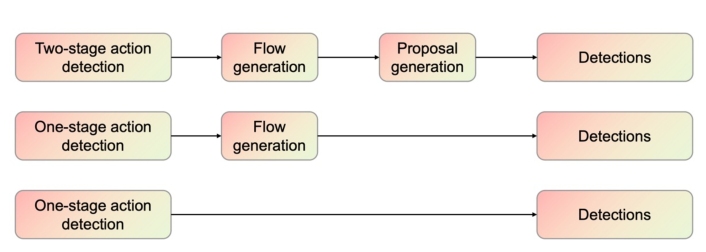}
\caption{ The image showcases three mainstream methods. the traditional two stream method, the two stream one stage method and the RGB only one stage method.} \label{fig2}
\end{figure}
In this paper, we present a novel TAD model called Mechanics-MLP, which employs a one-stage anchor-free framework and considers each token as a force. The Newtonian Mechanics-MLP unit inspired by MLP's success in computer vision backbones, achieved promising results using the $\beta-\text{GloU}$ loss for TAD, with our Tad-ML model achieving a maximum average precision (mAP) of 69.73\% (RGB and optical flow streams at tIoU=0.4) on THUMOS14 while exhibiting rapid processing speeds and superior accuracy compared to other methods, as demonstrated in Figure 1a. As shown in figure 2, illustrating that B represents a two-stage method, C represents a one-stage method, and D represents an end-to-end one-stage method. Our pro-posed method eliminates the need for the optical flow conversion pipeline, leading to faster processing. In conclusion, our Mechanics-MLP TAD model demonstrates promising results and offers a fresh perspective on designing one-stage anchor-free frameworks for TAD tasks. In summary, our paper has the following contributions:
{\begin{itemize}
\item[$\bullet$] {
To the best of our knowledge, TadML achieves state-of-the-art or highly competitive performance on benchmark data while significantly surpassing previous methods in terms of inference speed, achieving an impressive 4.44 videos per second inference speed on THUMOS14.}

\item[$\bullet$] {TadML demonstrates that optical flow data is not necessary for TAD tasks, thus improving the model's inference speed and improves performance in both RGB stream and two-stream by optimizing neck layers, while also finding $\beta-Giou$ to be a more appropriate metric for TAD.}

\item[$\bullet$] {Our Newtonian mechanics-based MLP model confirms the applicability of MLP for TAD and achieves highly competitive results using both RGB and optical flow data.}
\end{itemize}}

\section{Related Works}
This section provides a comprehensive review of previous studies related to Action Recognition, Temporal Action Detection, and MLP. 
\subsubsection{Temporal Action Recognition.}Action recognition in video clips, which involves identifying human actions in 2D frame sequences, has traditionally relied on manual feature extraction and classification through methods such as HOG, HOF, Dense Trajectories, SVM, and RF, but deep learning methods like Lrcns, R(2+1), and I3D have since become dominant in the field. These models use CNNs to extract spatial features, while RNNs like LSTM\cite{ref7} and 3D convolutions enable temporal feature extraction and improved re-mote loss and long-distance time modeling\cite{ref8}. The goal of sequential action detection is to identify action instances, time boundaries, and categories in videos, with two-stage methods dividing videos into proposals before assigning them to specific categories and one-stage models directly localizing and classifying actions for improved efficiency. The BMN model simplifies the BSN process and improves efficiency through a boundary matching mechanism, while PGCN\cite{ref9} employs graph convolutional networks to facilitate context and background information exchange. In contrast to two-stage models, one-stage models directly localize and classify actions, resulting in improved efficiency. For instance, SSAD simultaneously performs category prediction, time series offset correction, and IOU prediction, bypassing the requirement of initially predicting candidate time intervals. AFSD\cite{ref4} maximizes the utilization of boundary characteristics and extracts essential boundary features through boundary pooling. 
\subsubsection{MLP.} Various new computer vision architectures, including transformers and MLP, have recently demonstrated superior performance compared to CNN in several upstream tasks. Visual MLP-Like methods exhibit simplistic stacked MLP architectures, as exemplified by MLP-Mixer, which applies MLP independently and across image patches and has achieved comparable performance to SOTA models on the ImageNet dataset. The MorphMLP\cite{ref10} architecture emphasizes local information in the low-level layer and gradually transitions to a long-term model in the high-level layer. The WaveMLP\cite{ref11} introduce quantum mechanics into MLP. Both of these architectures have demonstrat-ed competitive performance in image classification tasks. Motivated by these findings, our objective is to investigate the potential of MLP-Like architectures in TAD, showcasing their applicability in visual downstream tasks as well.
\section{Method}
\subsection{Overview}TadML architecture features three elements, including a backbone module for feature extraction and down-sampling in time, a time fusion pyramid network (TFPN) as the neck, and action and time prediction branches acting as the head. Representing the untrimmed video datasets as $D = \{D_{train}, D_{test} \} $, each video in set as $ V\in\mathbb{R}^{T\times C\times H\times W}$, where T,C,H,W represent time step, channel, height and width respectively. In most TAD tasks, {V} will be converted into ${(\emph{V}_{rgb},\emph{V}_{opt})}$ first, where ${\textit{V}_{rgb}}$ contains RGB streams, and ${\textit{V}_{opt}}$ contains optical flow. This conversion takes a lot of time and calculation resources. Our model only takes RGB data as input. We obtain the output via the module, where output $Y =  {(d_{i,s}, d_{i,e}, c_i)}$. Here, $d_{i,s}$, $d_{i,e}$ are the distances between the current time step and the start and end of this action. A moment is either part of one action category or part of the background category, denoted with $c_i$.
\subsection{Architecture}
Mechanics token mixing block. Compared with the time complexity of the multi-head attention mechanism in Transformer, we aim to develop a similar approach that is relatively simple for rapid application in video understanding tasks. The MLP-like model is a neural architecture that is primarily made up of fully-connected layers and non-linear activation functions. We enhance token aggregation by dynamically adjusting the relationship between tokens and fixed weights in MLP through the application of Newtonian mechanics principles. In this work, given $D^0$ $\in \{X_1,X_2...,X_n\}$ with n time steps. $z^0$ is projected by  $F_a$ and $F_b$ with FC layer, where the angle between $F_a$  and $F_B$  is $\theta$ and $W^i$ and $W^j$ are the weight with learnable parameters. According to the laws of mechanics, their resultant force is $A_f$ , which is calculated by summing the vectors of $F_a$ and $F_b$.
 \begin{figure}
\centering
\includegraphics[width=0.4\textwidth]{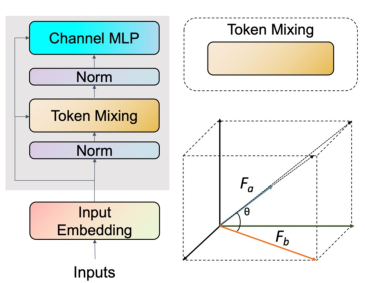}
\caption{Left the diagram of a block in the Mechanics-MLP architecture, right is token mixing.} \label{fig3}
\end{figure}
\begin{equation}
\resizebox{0.3\textwidth}{!}{$
\begin{aligned}
    &Token_{mechanics} = \{X_1,X_2...,X_n\}\\
    &F_a=FC(Token_{mechanics},W^i)\\
    &F_b=FC(Token_{mechanics},W^j)\\
    &A_f = \sqrt{F_a+F_b+2F_aF_b\cos\theta}
\end{aligned}
$}
\end{equation}
The inputs (embedding) in a basic mechanics unit undergo sequential processing through a mechanics token mixing block and a channel mixing block. Both two mixing block operation capture spatial information by blending features from multiple tokens. Furthermore, a layer normalization step is performed before each mixing operation, as depicted in Figure 3. The mechanics token mixing MLP consists of one MTM module, which aggregates various tokens by considering both $F_1$ and $F_2$, and applies the ReLU activation function. The channel mixing MLP extracts features for each token. 

\begin{figure}
\centering
\includegraphics[width=0.9\textwidth]{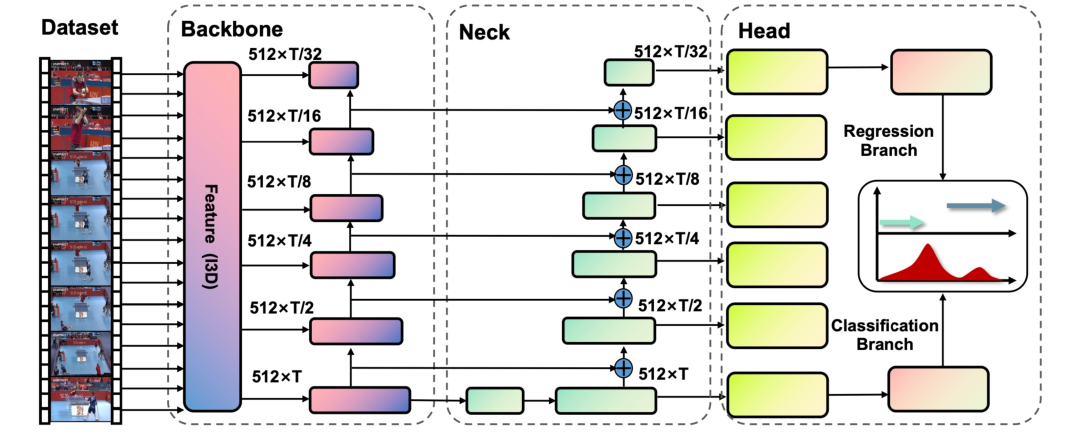}
\caption{The architecture consists of three main parts: a backbone module for feature extraction and downsampling in time, a time fusion pyramid network (TFPN) serving as the neck, and action and time prediction branches operating as the head.} \label{fig4}
\end{figure}
\begin{equation}
\resizebox{0.25\textwidth}{!}{$
\begin{aligned}
    &X = Norm(X)\\
    &Z = A_f(X)\\
    &Z = Norm(Z)\\
    &Z = Channel-FC(Z)
\end{aligned}
$}
\end{equation}
Backbone. To begin with, we generate video clips with a constant time window from the input video and reshape each clip to $(\emph{T},\emph{C},\emph{H},\emph{W})$. Feature extraction converts the video into a sequence of feature vectors corresponding to the RGB visual modality. Current TAD methods struggle to achieve fast detection due to their reliance on optical flow, which consumes computing resources and introduces time-consuming conversion processes. The conversion process is bypassed in our model's backbone module, eliminating the need for cumbersome operations. The backbone network utilizes a pre-trained I3D model on the Kinetics datasets to extract 3D features. The videos are segmented into short, overlapping 8-frame chunks. Finally, we obtain $Y_{I3D}=(Y_{rgb},Y_{opt})$, where both of them are 1024-dim features. Different from previous, our method only needs $Y_{rgb}$. We also tried $Y_{I3D}$ as our input, and it also played a good effect, indicating the superiority of our method. The feature is flattened along the last four dimensions to form a two-dimensional feature sequence encompassing both temporal and spatial information from the entire video. The feature sequence is then passed through a multi-layer semantic module (MSM) consisting of six down-sample layers. Each layer, composed of mechanics units, has an output dimension of 512. The outputs of the layers are (512, t/2), (512, t/4), (512, t/8), (512, t/16), (512, t/32), and (512, t/64), as shown in Figure 4. This module produces multi-coarse texture basic features as its output.
Temporal Feature Pyramid Network. the TAD architecture incorporates the concept of object detection, making the utilization of complex adaptive attention modules in AFFPN and the repeated superposition method in BIFPN impractical. The neck is designed to establish connections between semantically strong, low-temporal-resolution features and semantically weak, high-temporal-resolution features. This is achieved through a six-layer pathway comprising ((T,512), (T/2,512), (T/4,512), (T/8,512), (T/16,512), (T/32,512)). The six-layer design proves highly effective in extracting temporal content. The high-resolution features extracted by the backbone are sequentially up-sampled to each of the six layers. The up-sampling operation involves bi-linear interpolation and simultaneous combination with the high-resolution features. This modified approach simplifies the model while capturing features from more detailed examples.
Temporal Action Detection Heads. In TadML, the Temporal Action Detection Heads (TADH) simultaneously predict action categories and temporal boundaries through two branches: the classification branch, which estimates the probability of each class $c_i$ and the regression branch that forecasts the starting and ending time distance $(t_{i,s}, t_{i,e})$. Each time step t further decodes an action instance, including: action start time and action end time denoted by $(T_{s} = T-T^e$ and $T_{e} = T+T^s)$ and an action confidence score denoted with $c_i$. Both branches are constructed with three MLP layers and two LayerNorm layers. An additional activation ReLU layer is included in the classification branch to predict the category ID.

Loss Construction. $\mathcal{L}_{cls}$  is Focal loss that employed as the classification loss, effectively mitigating class imbalance by adjusting the weights of positive and negative samples based on their classification difficulty levels, thereby enhancing overall detection accuracy. $\mathcal{L}_{reg}$  is the regression loss that defined to differentiate the regression of instance time boundaries. For the regression loss, we define it to distinguish the instance time boundary regression. We propose an improvement of GIoU, called $\beta-GIoU$ , to constructed the regression loss. In $\beta-GIoU$, the hyper-parameter $\beta$ is defined to detect shapes sensitively in the error term of the predicted value and position, A is candidate object, the B represents the object ground-truth and C is minimum bounding box area. To accomplish this, the total loss is defined by two super parameters $\lambda_{cls}$ and $\lambda_{reg}$ are set up for classification loss and regression loss separately. $T_{at}$ is an indicator function indicating to determine whether there is an action in the time step. while $T_{+}$ represents the total number of positive samples.
\begin{equation}
\resizebox{0.4\textwidth}{!}{$
\begin{aligned}
    &\mathcal{L}_{reg}=\mathcal{L}_{\beta-\text { GloU }}=1-IoU+\left(\frac{\left|C \backslash\left(A \cup B^{g t}\right)\right|}{|C|}\right)^{\beta}\\
    &\mathcal{L}_{cls}\left(\left\{\hat{y}_{i}\right\}\right)=\frac{1}{N} \sum_{i} \ell_{\text {focal }}\left(\hat{y}_{i}, y_{i}\right)\\
    &\mathcal{L} = \sum_{k=1}^N (\frac{\mathcal{L}_{cls}}{T}\ell_{cls}+ \frac{\lambda_{reg}}{T_+}\mathbb{T}_{at}\mathcal{L}_{reg})
\end{aligned}
$}
\end{equation}

Train and Inference. We implement our network using the pytorch framework. All experiments were run on a workstation equipped a single Tesla P100 GPU, and Intel(R) Xeon(R) CPU (E5-2690 v4 @ 2.90GHz). The models were trained for 80 epochs using Adam with warm-up for training, which is crucial for achieving model convergence and optimal performance. The base learning rate is set to $10^{-5}$ and the batch size was set to 4. The weights of the loss terms, $\lambda_{cls}$ and $\lambda_{reg}$ were both set to 1. The parameter $\beta$, in $\beta-Giou$, was set to 3. Input sequences were cropped or padded to the maximum length 2304. During the inference process, only the action predictions from the last lightweight MLP layer are considered, while the full sequences are fed into the model. Our model takes the input video X and outputs $(t_{i,s},t_{i,e},c_{i})$ for each time step T across all neck levels. The final TAD results are generated by processing action candidates with Non-maximum Suppression (Soft-NMS) to remove highly overlapping instances.

\section{Experiments and Results}
We evaluated the effectiveness of our proposed approach through benchmark evalua-tions on THUMOS14\cite{ref12} and ActivityNet1.3\cite{ref13}, as well as extensive ablation stud-ies to analyze model performance.

\subsection{Evaluation}
For all datasets, we report the standard mean average precision (mAP) at different temporal intersection over union (tIoU) thresholds, which is widely used to evaluate TAD models. For the THUMOS14, the tIoU threshold is selected from {0.3, 0.4, 0.5, 0.6, 0.7}. For ActivityNe-t1.3, the tIoU threshold is {0.5, 0.75, 0.95}. We also report the average graph with fine-scale tIoU threshold ([0.5, 0.95] step is 0.05). 
\begin{table}
\caption{Performance comparison with  methods on THUMOS14, measured by mAP at different IoU thresholds, and average mAP in [0.3 : 0.1 : 0.7] on THUMOS14.}\label{tab1}
\centering
  \resizebox{0.45\textwidth}{!}
  {\begin{tabular}{|c@{}|l|c|c|c|c|c|c|c@{}|}
\toprule
Type& Method& RGB stream&\quad0.3&0.4&0.5&0.6&0.7&Avg\\
\hline
  \multirow{2}*{Two-stage} 
 &CDC\cite{ref14} &\tikz \node[cross out,draw=black,thick] {}; &\quad40.10& 29.40& 23.30& 13.10& 7.90& 20.76\\
 &TCN\cite{ref15}&\tikz \node[cross out,draw=black,thick] {}; &\quad-& 33.30& 25.60& 15.90& 9.00&-\\
 &TURN-TAP\cite{ref16}&\tikz \node[cross out,draw=black,thick] {}; &\quad44.10& 34.90& 25.60& -& -&-\\
 &R-C3D\cite{ref17}&\tikz \node[cross out,draw=black,thick] {}; &\quad44.80& 35.60& 28.90& -& -&-\\
 &MGG\cite{ref18} &\tikz \node[cross out,draw=black,thick] {};  &\quad53.9&46.8&37.4&29.5&21.3& 37.78\\
 &BMN\cite{ref3} &\tikz \node[cross out,draw=black,thick] {}; &\quad56&47.4&38.8&29.7&20.5& 38.48\\
 &DBG\cite{ref19}&\tikz \node[cross out,draw=black,thick] {}; &\quad57.8& 49.4& 39.8& 30.2& 21.7&39.78\\
 &BSN++\cite{ref20}&\tikz \node[cross out,draw=black,thick] {}; &\quad59.90& 45.90& 41.30& 31.90& 22.80&40.36\\
 &GCN\cite{ref9}&\tikz \node[cross out,draw=black,thick] {}; &\quad63.6& 57.8& 49.1& -& -&-\\
 &TAL-Net\cite{ref21}&\tikz \node[cross out,draw=black,thick] {}; &\quad53.2&48.5&42.8&33.8&20.8&39.8 \\
 &G-TAD\cite{ref22}&\tikz \node[cross out,draw=black,thick] {}; &\quad58.7& 52.7& 44.9& 33.6& 23.8& 42.7\\
 &MR\cite{ref23}&\tikz \node[cross out,draw=black,thick] {}; &\quad53.9& 50.7& 45.4& 38.0& 28.5& 43.3\\
 &ContextLoc\cite{ref24}&\tikz \node[cross out,draw=black,thick] {}; &\quad68.3& 63.8& 54.3& 41.8& 26.2&50.88\\
\hline
\multirow{2}*{One-stage} & PBRNet\cite{ref25} &\tikz \node[cross out,draw=black,thick] {};  &\quad58.5& 54.6& 51.3& 41.8& 29.5& -\\
&A2Net\cite{ref26} &\tikz \node[cross out,draw=black,thick] {};  &\quad58.6& 54.1& 45.5& 32.5& 17.2& 41.6\\
&A2Net &\XSolid &\quad58.6& 54.1& 45.5& 32.5& 17.2& 41.6\\
&G-TAD\cite{ref27}& \Checkmark &\quad57.8&47.2&38.8&-&-&-\\
&TadTR\cite{ref1} &\tikz \node[cross out,draw=black,thick] {};  &\quad62.4&57.4&49.2&37.8&26.3&46.6 \\
&TadML & \Checkmark &\quad68.78&64.66&56.61&45.40&31.88&53.46\\
&TadML &\tikz \node[cross out,draw=black,thick] {}; &\quad73.29&69.73&62.53&53.36&39.60&59.70\\
\hline
\end{tabular}
}
\end{table}
THUMOS14 is comprised of 413 untrimmed videos spanning 20 action categories. Each video contains 15 action instances on average, each instance has an average of 8\% overlapping with others. The datasets are divided into two subsets: a verification set and a test set. The verification set contains 200 videos and the test set contains 213 videos. Following the standard setup, we use validation sets for training and the testing videos for evaluation. The experimental results on THUMOS14 are shown in table 1. The results are presented in Table 1. Without optical flow input, our method achieves an average mAP of 53.46\% ([0.3: 0.1: 0.7]), with a mAP of 56.61\% at tIoU=0.5 and a mAP of 31.88\% at tIoU=0.7. This result exceeds those of most methods of TAD, even including models with additional optical flow input. This suggests that our model is not only achieves comparable in accuracy but also faster than most methods in practice. This is because our model skips the step of conversion (from raw RGB to optical flow), which is especially time-consuming. In order to further prove the superiority of our model, we also conducted experiments using optical flow input. The results show that our model achieves an average mAP of 59.7\% ([0.3 : 0.1 : 0.7]), with a mAP of 62.53\% at tIoU=0.5 and a mAP of 39.6\% at tIoU=0.7.
\begin{table}
\caption{Performance comparison with methods on ActivityNetv1.3, measured by mAP at different IoU thresholds, and average mAP in [0.5 : 0.75 : 0.95] on ActivityNetv1.3.}\label{tab2}
\centering
  \resizebox{0.45\textwidth}{!}
  {\begin{tabular}{|c@{}|l|c|c|c|c|c|c|c@{}|}
\hline
 Method&Single-stage&\quad0.5&0.75&0.95& Avg\\
\hline
R-C3D\cite{ref17}&\XSolid&\quad26.80&\quad$-$&\quad$-$&\quad$-$\\
TAL-Net\cite{ref21}&\XSolid&\quad38.23&\quad18.30&\quad1.30&\quad20.22\\
BSN\cite{ref2}&\XSolid&\quad56.45&\quad29.96&\quad8.02&\quad30.03\\
BMN \cite{ref3}&\XSolid&\quad50.07&\quad34.78&\quad8.29&\quad33.85\\
P-GCN\cite{ref9}&\XSolid&\quad42.90&\quad28.14&\quad2.47&\quad26.99\\
Contextloc\cite{ref24}&\XSolid&\quad51.24&\quad 31.40&\quad 2.83&\quad 30.59 \\
TadTR+BMN\cite{ref1}&\XSolid&\quad50.51&\quad 35.35&\quad 8.18&\quad 34.55\\
A2Net\cite{ref26}&\Checkmark&\quad43.55&\quad 28.69&\quad 3.70&\quad 27.75\\
SSN\cite{ref28}&\Checkmark&\quad43.26&\quad28.70&\quad5.63&\quad28.28\\
TadTR\cite{ref1}&\Checkmark&\quad49.08&\quad 32.58&\quad 8.49&\quad 32.27\\
G-TAD\cite{ref27}&\Checkmark&\quad50.36&\quad 34.60&\quad 9.02&\quad 34.09\\
AFSD\cite{ref4}& \Checkmark&\quad52.4&\quad 35.3&\quad 6.5&\quad 34.4\\
Ours  &\Checkmark&\quad53.15&\quad35.75&\quad7.47&\quad34.94\\
\hline
\end{tabular}
}
\end{table}
ActivityNet1.3 is a large-scale action dataset that comprises 200 action classes and approximately 2k untrimmed videos. And it's total video length exceeds 600 hours. The dataset has been split into three subsets, with 10,024 videos for training, 4,926 videos for validation, and 5,044 videos for testing. In line with the established meth-odology, we trained TadML on the training set and test, the performance on the validation set. The experiment results on ActivityNet v1.3 are presents in table 2. The results are presented in Table 2. Using I3D features, our method achieves an average mAP of 34.94\% ([0.5 : 0.05 : 0.95]), this outperforms all previous methods that use the same features by at least 0.6\%. This improvement is significant as it is averaged across multiple tIoU thresholds, including those tight ones e.g. 0.95. Furthermore, by employing the pre-training method from TSP, we slightly improve our results, achieving an 36.0\% average mAP. Our model thus outperforms the best method with the same features by a small margin. Again, our method largely outperforms TadTR. Our results are only inferior to TCANet—a latest two-stage method using stronger SlowFast features. We conjecture that our method will also benefit from better features. Nevertheless, our simple model clearly demonstrates state-of-the-art results on this challenging dataset.
\subsection{Ablation Study}
In this section, we conducted several ablation studies on THUMOS14 to further verify the efficacy of our model. Our experiments examined the efficacy of key components and recommended hyper-parameters settings. For all experiments, we kept the evaluation settings constant and only made changes to the corresponding components. 
\begin{table}
\caption{Study of different number of frozen stages of backbone
on THUMOS14 in terms of mAP(\%)@tIoU.}\label{tab3}
\centering
  \resizebox{0.45\textwidth}{!}
{\begin{tabular}{|c|c|c|c|c|c|c|}
\hline
 Neck Stages&RGB&\quad0.3&0.4&0.5&0.6&0.7\\
\hline
1&\XSolid&\quad56.09&49.11&38.35&24.92&12.03 \\
2&\XSolid&\quad61.74&55.6&44.81&29.39&14.29  \\
3&\XSolid&\quad65.74&60.16&49.91&36.09&19.78  \\
4&\XSolid&\quad66.82&62.32&53.82&43.26&28.96  \\
5&\XSolid&\quad66.98&62.77&55.42&44.81&31.82  \\
6&\XSolid&\quad68.7&64.66&56.61&45.40&31.88 \\
7&\XSolid&\quad68.7&64.66&56.61&45.40&- \\
1&\Checkmark& \quad62.7&57.06&46.64&30.73&14.52\\
2&\Checkmark& \quad68.04&62.6&52.37&35.65&18.52\\
3&\Checkmark& \quad70.78&66.1&57.6&44.36&27.41\\
4&\Checkmark& \quad68.7&64.66&56.61&45.40&-\\
5&\Checkmark& \quad73.32&68.91&62.28&52.81&39.04\\
6&\Checkmark& \quad72.79&69.49&62.72&52.29&38.94\\
7&\Checkmark& \quad73.59&69.69&62.79&53.13&40.22\\
\hline
\end{tabular}
}
\end{table}
Neck layers play a crucial role in temporal action detection, as evidenced by our comparison of different neck layers and RGB streams presented in Table 3. The number of neck layers ranges from 1 to 7, and as the number of neck layers increases, the average mAP also increased from 36.03\% to 53.46\%, and reach its peak at 6th neck layer. At this point, when the neck layer added again, the performance stars to decline. Furthermore, our method also exhibits great performance in two-streams. with the number of layers are set to 7, achieving an average mAP is 59.7\%. We have also conducted comparisons using different MLP-Like blocks. While keeping other parameter settings keep unchanged. The results are presented in Table 4, when only RGB streams are used, the average MAP achieved with Waveblock is 51.64\%, with MorphMLP it is 52.09\%, and with TadML it is 53.46\%. To further evaluate our model, we have also included optical flow for comparison.  Our model achieved an average MAP as high as 59.70\%. These results are shown in Table 5. When the weight of classification and regression loss is set at 1, the best performance is achieved by $\beta$, which achieves a MAP of 53.46\% when the value of $\beta$ is 3.
\begin{table}
\caption{Study of three different backbone (WaveMLP, MorphMLP, Mechaincs-MLP)
on THUMOS14 in terms of mAP(\%)@tIoU.}\label{tab4}
  \centering
  \resizebox{0.45\textwidth}{!}{
    \begin{tabular}{|l|c|c| c| c| c| c |c|}
    \hline
    Neck Stages&RGB&\quad0.3&0.4&0.5&0.6&0.7&Avg\\
    \hline
    WaveMLP\cite{ref11}&\XSolid&\quad66.87&62.46&54.33&44.00&30.53&51.64 \\
    WaveMLP\cite{ref11}&\Checkmark&\quad72.01&68.02&61.51&52.01&38.28&58.36\\
    MorphMLP\cite{ref10}&\XSolid&\quad66.91&62.83&54.93&44.57&31.20&52.09 \\
    MorphMLP\cite{ref10}&\Checkmark&\quad72.21&69.12&62.87&52.55&38.47&59.04\\
    Ours  & \XSolid&\quad68.78&64.66&56.61&45.40&31.88&53.46\\
    Ours  &\Checkmark&\quad73.29&69.73&62.53&53.36&39.60&59.70\\
    \hline
\end{tabular}
}
\end{table}
\begin{table}
\caption{Study of different $\beta$ ($\beta-Giou$) on THUMOS14 in terms of mAP(\%)@tIoU.}\label{tab5}
  \centering
  \resizebox{0.45\textwidth}{!}{
    \begin{tabular}{|c|c| c| c |c |c |c|}
      \hline
      $\beta$&\quad0.3&0.4&0.5&0.6&0.7&Avg\\
      \hline
      1&\quad66.68&63.60&56.55&43.77&31.24&52.57 \\
      2&\quad67.95&63.55&56.68&45.12&31.97&53.05\\
      3&\quad68.78&64.66&56.61&45.40&31.88&53.46\\
      4&\quad67.75&64.03&56.47&43.74&31.34&52.67\\
      5&\quad67.64&63.82&56.43&44.06&31.17&52.63\\
      10&\quad67.44&63.80&56.42&44.15&30.81&52.52\\
      \hline
    \end{tabular}
  }
\end{table}
\section{Conclusion}
In this work, we introduce TadML, an anchor-free one-stage MLP method designed for TAD using RGB stream input. Our method simplifies the traditional TAD pipe-line by eliminating the need for manual conversion of optical flow data. Additionally, we propose $\beta-GloU$ for the framework. To the best of our knowledge, TadML is the first MLP-like model suitable for TAD. We leverage Newtonian Mechanics to address the token mixing problem. TadML showcases the potential of MLP-like methods in downstream visual tasks, surpassing many recent methods (including those using optical flow input) and offering twice the inference speed of BMN. Due to its independence from optical flow conversion, our method holds promise for practical applications in the field of TAD. Moreover, it achieves impressive performance when both RGB and flow data are utilized. Our goal is to promote the development of efficient models for temporal action detection and facilitate their adoption in industrial set-tings. The source code is available at \url{https://github.com/BonedDeng/TadML}.

\subsubsection{Acknowledgements} Thanks for the support of supervisor Zhao in Guangxi University and the internship opportunities provided by supervisor Liu in the Institute of Automation, Chinese Academy of Sciences.

%
%
%
%

\end{document}